\documentclass{article} 
\pdfoutput=1
\usepackage{nips13submit_e,times}
\usepackage{hyperref}
\usepackage{url}
\usepackage{paralist}
\usepackage{multirow}
\usepackage{verbatim}
\usepackage{graphicx}
\usepackage{cleveref}

\graphicspath{ {images/} }

\def\note#1{\textcolor{red}{{\texttt{NOTE:}}} \texttt{#1}}

\title{AtomNet: A Deep Convolutional Neural Network for Bioactivity Prediction in Structure-based Drug Discovery}

\author{
Izhar Wallach\\
Atomwise, Inc.\\
\texttt{izhar@atomwise.com} \\
\And
Michael Dzamba\\
Atomwise, Inc.\\
\texttt{misko@atomwise.com} \\
\And
Abraham Heifets\\
Atomwise, Inc.\\
\texttt{abe@atomwise.com} \\
}

\nipsfinalcopy

\begin{document}

\maketitle

\begin{abstract}
Deep convolutional neural networks comprise a subclass of deep neural networks (DNN) with a constrained architecture that leverages the spatial and temporal structure of the domain they model. 
Convolutional networks achieve the best predictive performance in areas such as speech and image recognition by hierarchically composing simple local features into complex models. 
Although DNNs have been used in drug discovery for QSAR and ligand-based bioactivity predictions, none of these models have benefited from this powerful convolutional architecture. 
This paper introduces AtomNet, the first structure-based, deep convolutional neural network designed to predict the bioactivity of small molecules for drug discovery applications.  
We demonstrate how to apply the convolutional concepts of feature locality and hierarchical composition to the modeling of bioactivity and chemical interactions.
In further contrast to existing DNN techniques, we show that AtomNet's application of local convolutional filters to structural target information successfully predicts new active molecules for targets with no previously known modulators.  
Finally, we show that AtomNet outperforms previous docking approaches on a diverse set of benchmarks by a large margin, achieving an AUC greater than 0.9 on 57.8\% of the targets in the DUDE benchmark. 

\end{abstract}

\section{Introduction}
Fundamentally, biological systems operate through the physical interaction of molecules. 
The ability to determine when molecular binding occurs is therefore critical for the discovery of new medicines and for furthering of our understanding of biology. 
Unfortunately, despite thirty years of computational efforts, computer tools remain too inaccurate for routine binding prediction, and physical experiments remain the state of the art for binding determination. 
The ability to accurately predict molecular binding would reduce the time-to-discovery of new treatments, help eliminate toxic molecules early in development, and guide medicinal chemistry efforts~\cite{Kitchen2004,Lingle2015}.

In this paper, we introduce a new predictive architecture, AtomNet, to help address these challenges. 
AtomNet is novel in two regards: 
AtomNet is the first deep convolutional neural network for molecular binding affinity prediction. 
It is also the first deep learning system that incorporates structural information about the target to make its predictions. 

Deep convolutional neural networks (DCNN) are currently the best performing predictive models for speech and vision~\cite{LeCun2015, Krizhevsky2012, Hinton2012, Szegedy2014}. 
DCNN is a class of deep neural network that constrains its model architecture to leverage the spatial and temporal structure of its domain. 
For example, a low-level image feature, such as an edge, can be described within a small spatially-proximate patch of pixels. 
Such a feature detector can share evidence across the entire receptive field by ``tying the weights" of the detector neurons, as the recognition of the edge does not depend on where it is found within an image~\cite{LeCun2015}.
This reduction in the number of model parameters reduces overfitting and improves the discovery of generalizable features. 
Local low-level features are then hierarchically composed by the network into larger, more complex features ({\em e.g.}, for a face recognition task, pixels may be combined into edges; edges into eyes and noses; eyes and noses into faces)~\cite{Yosinski2015}. 

Our insight is that biochemical interactions are similarly local, and should be modeled by similarly-constrained machine learning architectures. 
Chemical groups are defined by the spatial arrangement and bonding of multiple of atoms in space, but these atoms are proximate to each other.
When chemical groups interact, {\em e.g.} through hydrogen bonding or $\pi$-bond stacking, the strength of their repulsion or attraction may vary with their type, distance, and angle, but these are predominantly local effects~\cite{Bissantz2010}. 
More complex bioactivity features may be described by considering neighboring groups that strengthen or attenuate a given interaction but, because even in these cases distant atoms rarely affect each other, the enforced locality of a DCNN is appropriate. 
Additionally, as with edge detectors in DCNNs for images, the applicability of a detector for {\em e.g.}, hydrogen bonding or $\pi$-bond stacking, is invariant across the receptive field. 
These local biochemical interaction detectors may then be hierarchically composed into more intricate features describing the complex and nonlinear phenomenon of molecular binding. 

In addition to introducing the DCNN architecture for biochemical feature discovery, AtomNet is the first deep neural network for {\em structure-based} binding affinity prediction. 

Recently, deep neural networks have been shown to out-perform random forests and SVMs for QSAR and ligand-based virtual screening~\cite{Dahl2014,Junshui2015,Unterthiner2015}. 
Introduced by Dahl~{\em et~al.}~\cite{Dahl2014}, the best performing architecture for the Merck Molecular Activity Kaggle Challenge~\cite{Junshui2015} was a multi-task deep neural network (MT-DNN). 
The multi-task architecture trains a single neural network with multiple output neurons, each of which predict the activity of the input molecule in a different assay. 
Because molecules are often tested in multiple assays, the MT-DNN architecture can combine training evidence among similar prediction tasks~\cite{Dahl2014}.
That work was followed by Untherhiner~{\em et~al.}~\cite{Unterthiner2015,Unterthiner2015-2} and Ramsundar~{\em et~al.}~\cite{Ramsundar2015} that demonstrated the MT-DNN technique scales to large biochemical databases such as PubChem Bioassays~\cite{Wang2012} and ChEMBL~\cite{Bento2014}. 

Ligand-based techniques, including MT-DNN, come with several limitations. 
First, they are restricted to targets for which substantial amounts of prior data are already available and, as such, cannot make predictions for novel targets. 
In practice, this creates a paradoxical dynamic -- these predictive models offer the most help precisely for those targets which least require it. 
The dependence on known active ligands also makes it difficult to show that the network is ``right for the right reasons"; artifacts in the training data, such as analogue bias, make it very difficult to properly assess accuracy and generalizability~\cite{Rohrer2009,Xia2014,Lagarde2015}. 
Second, existing deep neural networks for ligand-based models take molecular fingerprints, such as ECFP~\cite{David2010}, as input. 
Such input encoding limits the discovery of features to compositions of the pre-specified molecular structures defined during the fingerprinting process~\cite{Unterthiner2015} and eliminates the ability to discover arbitrary features. 
Third, as the model is blind to the target, the model cannot elucidate which potential interactions are left unfulfilled by a molecule. 
This limits the guidance that could be provided to medicinal chemists for optimization of the molecule. 

To address these limitations, AtomNet combines information about the ligand with information about the structure of the target. 
Our approach requires the locations of each atom in the binding site of the target (a burden that ligand-based approaches avoid), but access to this information enables the model to discover arbitrary molecular features. 
These features describe favorable and unfavorable interactions between ligands and targets and, as shown in Section~\ref{sec:Results}, can be applied to targets for which no binders are known by the model. 

In the following, we present the design and development of AtomNet and report its performance on a range of challenging structure-based bioactivity prediction experiments.

\section{Methods}
\label{sec:Methods}

We first describe the construction of the experimental benchmarks on which we test our system. 
Then we describe our data encoding and the design of our deep convolutional network.

\subsection{Datasets}
We demonstrate the application of our AtomNet model on three realistic and challenging benchmarks: the Directory of Useful Decoys Enhanced (DUDE) benchmark~\cite{Mysinger2012}; our internal DUDE-like benchmark; and a benchmark with experimentally-verified inactive molecules. 
Each of these benchmarks provide a different and complimentary assessment of our performance; the advantages of each are summarized here, and described in more detail below. 
As the standard benchmark, DUDE permits direct comparisons to other structure-based binding affinity prediction systems. 
Unfortunately, DUDE only specifies a test set, without specifying a separate training set; by constructing our own DUDE-like benchmark, we can ensure that there is no overlap between the training and test molecules.
Finally, correctly classifying experimentally-verified active and inactive molecules is a challenging test because structurally similar molecules can have different labels~\cite{Hu2012}. 
Such cases are excluded from benchmarks using property-matched decoys because of the dissimilarity requirement in order to presume decoys are inactive.

\paragraph{DUDE}
The DUDE is a well-known benchmark for structure-based virtual screening methods from the Shoichet Lab at UCSF~\cite{Mysinger2012}. 
The methodology of the DUDE benchmark is fully described by Mysinger~{\em et~al.}~\cite{Mysinger2012}. 
Briefly, the benchmark is constructed by first gathering diverse sets of active molecules for a set of target proteins. 
Analogue bias is mitigated by removing similar actives; similar actives are eliminated by first clustering the actives based on scaffold similarity, then selecting exemplar actives from each cluster.
Then, each active molecule is paired with a set of property matched decoys (PMD)~\cite{Wallach2011}. 
PMD are selected to be similar to each other and to known actives with respect to some 1-dimensional physico-chemical descriptors (\textit{e.g.,} molecular weight) while being topologically dissimilar based on some 2D fingerprints (\textit{e.g.,} ECFP~\cite{David2010}). 
The enforcement of the topological dissimilarity supports the assumption that the decoys are likely to be inactives because they are chemically different from any know active.
The benchmark consists of 102 targets, 22,886 actives (an average of 224 actives per target) and 50 PMD per active. 
We randomly selected 30 targets as our test set and designated the remaining 72 targets as the training set. 

\paragraph{ChEMBL-20 PMD}
We constructed a DUDE-like dataset derived from ChEMBL version 20~\cite{Bento2014}. 
We considered all activity measurements that passed the following filters:
\begin{compactitem}
\item Affinity units measured in IC50 or Ki and lower than $1\mu M$.
\item Target confidence greater or equal to 6.
\item Target has an annotated binding site in the scPDB database~\cite{Desaphy02014} and resolution $<2.5$\AA.
\item Ligands passed PAINS filers~\cite{Baell2010} and promiscuity rules~\cite{Bruns2012}.
\end{compactitem}

Following Mysinger~{\em et~al.}~\cite{Mysinger2012}, we first grouped target affinities by their UniProt gene name prefix~\cite{UniProt2015} and removed targets for which there were less than 10 active ligands. 
This filtering process yielded a set of 123,102 actives and 348 targets. 
Second, each active was paired with a set of 30 PMD selected from the ZINC database~\cite{Irwin2005} similarly to Mysinger~{\em et~al.}~\cite{Mysinger2012}.
Third, we partitioned the data into training, validation, and testing sets by first clustering the active ligands for each target based on their Bemis-Murcko scaffolds~\cite{Bemis1996} and choosing ligands that were at least $3\mu M$ apart as the cluster exemplars.
Clusters with less than 10 exemplars were discarded.
Fourth, we defined the test set by randomly selecting 50 targets with their corresponding actives and decoys.
Last, the training set was further partitioned over the clusters into 5-fold cross validation sets.
The final dataset consists of 78,904 actives, 2,367,120 decoys, and 290 targets. 

\paragraph{Experimentally verified inactives}
A limitation of benchmarks based on PMD is that they exclude decoys that are similar to active molecules. 
This design decision is to justify the assumption that selected decoys are likely to be inactive, even without experimental validation. 
This enforced dissimilarity between actives and decoys means that PMD benchmarks lack some challenging cases where actives and inactive molecules are highly similar~\cite{Hu2012}.

We include these challenging cases by substituting decoys with molecules that have been experimentally validated to be inactives. 
We constructed a benchmark similar to our ChEMBL-20 PMD one but replaced PMD with inactive molecules. 
We defined a molecule as inactive if its measured activity is higher than $30\mu M$. 
We ended up with a set of 78,904 actives, 363,187 inactives, and 290 targets which was partitioned into 3-fold cross-validation sets over the Bemis-Murcko clusters. 
Targets with less than 10 clusters were never assigned into a validation set.
Hence, the number of targets in the validation sets was 149.

\subsection{Structure-based deep-convolutional neural network}
The network topology consists of an input layer, followed by multiple 3D-convolutional and fully-connected layers, and topped by a logistic-cost layer that assigns probabilities over the active and inactive classes. 
All units in hidden layers are implemented with the ReLU activation function~\cite{Nair2010}. 

\paragraph{Input representation}
The input layer receives vectorized versions of 1\r{A} 3D grids placed over co-complexes of the target proteins and small-molecules that are sampled within the target's binding site. 
First, we define the binding site using a flooding algorithm~\cite{Hendlich1997} seeded by a bound ligand annotated in the scPDB database~\cite{Desaphy02014}.
Second, we shift the coordinate of the co-complexes to a 3D Cartesian system originated at the center-of-mass of the binding site.
Third, we sample multiple poses within the binding site cavity.
Fourth, we crop the geometric data to fit within an appropriate bounding box. 
In this study we used a cube of 20\r{A}, centered at the origin.
Fifth, we translate the input data into a fixed-size grid with 1\r{A} spacing.
Each grid cell holds a value that represents the presence of some basic structural features in that location. 
Basic structural features can vary from a simple enumeration of atom types to more complex protein-ligand descriptors such as SPLIF~\cite{Da2014}, SIFt~\cite{Deng2004}, or APIF~\cite{Nueno2009}.
Last, we unfold the 3D grid into a 1D floating point vector. 

\paragraph{Network architecture}
3D convolutional layers were implemented to support parameters such as filter size, stride, and padding in a similar fashion to the implementation of Krizhevsky~{\em et~al.}~\cite{Krizhevsky2012}. 
We used network architecture of an input layer as described above, followed by four convolutional layers of  $128\times5^3$, $256\times3^3$, $256\times3^3$, $256\times3^3$ (number of filters $\times$ filer-dimension), and two fully-connected layers with 1024 hidden units each, topped by a logistic-regression cost layer over two activity classes. 

\paragraph{Model Training}
Training the model was done using stochastic gradient descent with the AdaDelta adaptive learning method~\cite{Zeiler2012}, the backpropagation algorithm~\cite{Rumelhart1988}, and mini-batches of 768 examples per gradient step. 
No attempt was made to optimize meta-parameters except the limitation of fitting the model into a GPU memory.
Training time was about a week on 6 Nvidia-K10 GPUs.

\paragraph{Baseline method for comparison}
We used Smina~\cite{Koes2013}, a fork of AutoDock Vina~\cite{Trott2010}, as a baseline for the SB evaluation. 
Smina implements an improved empirical scoring function and minimization routines over its predecessor and is freely available under the GPLv2 license. 

\section{Results}
\label{sec:Results}
We use the area under the receiver operating characteristic (AUC) and logAUC to report results over the three benchmarks. 
The AUC indicates classification (or ranked order) performance by measuring the area under the curve of true-positive rate versus the false-positive rate. 
AUC value of 1.0 means perfect separation whereas a value of 0.5 implies random separation.
LogAUC is a measurement similar to AUC that emphasizes early enrichment performance by putting more weight at the beginning of the curve so cases correctly classified at the top of the rank-ordered list contribute more to the score than later ones.
Here, we used logarithmic base of 10 which means that the weight of the first 1\% of the ranked results equal to the weight of the next 10\%.
Because the non-linearity of a logAUC value makes it hard to interpret, we subtract the area under the log-scaled random curve (0.14462) from a logAUC to get an adjusted-logAUC~\cite{Mysinger2010}.
Hence, positive adjusted-logAUC values imply better than random performance whereas negative ones imply worse than random performance. 
For brevity, we will use adjusted-logAUC and logAUC interchangeably for the rest of this manuscript.

\begin{figure}[!htpb]
  \begin{center}
    \includegraphics[width=\textwidth]{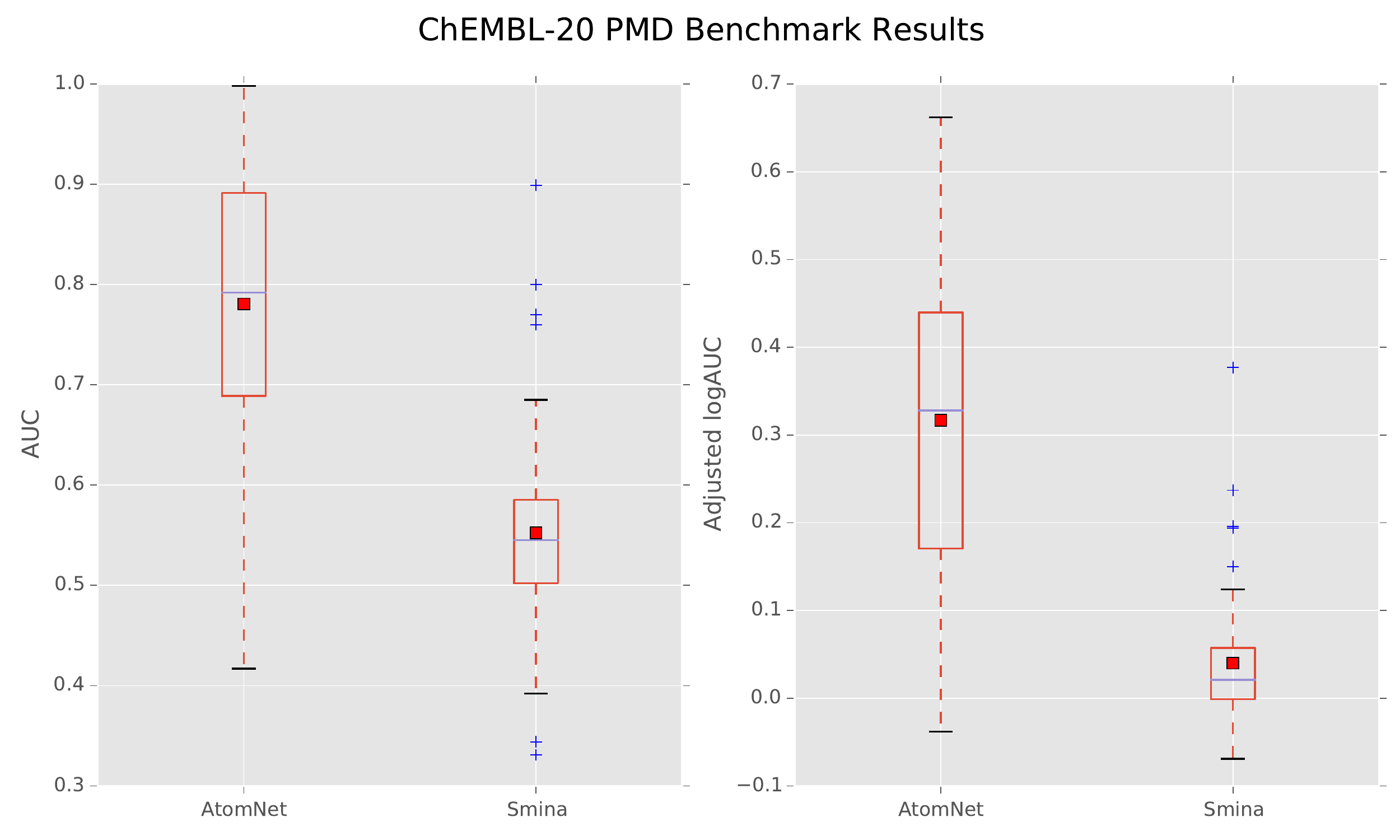}
  \end{center}
  \caption{Distribution of AUC and logAUC values of 50 ChEMBL-20-PMD targets for AtomNet and Smina.}
  \label{fig:chemblPMD}
\end{figure}

\begin{figure}[!htpb]
  \begin{center}
    \includegraphics[width=\textwidth]{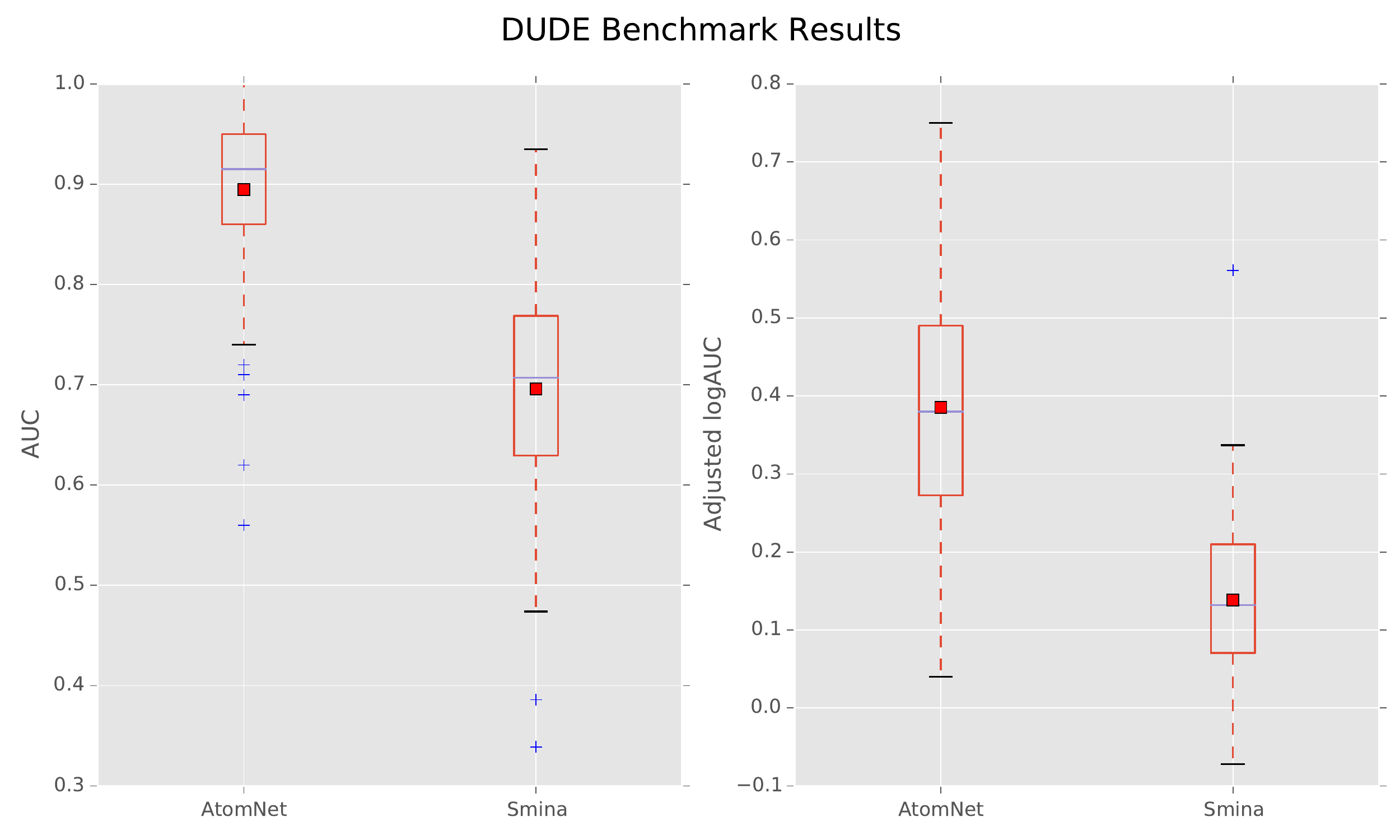}
  \end{center}
  \caption{Distribution of AUC and logAUC values of 102 DUDE targets for AtomNet and Smina.}
  \label{fig:DUDE}
\end{figure}

\begin{figure}[!htpb]
  \begin{center}
    \includegraphics[width=\textwidth]{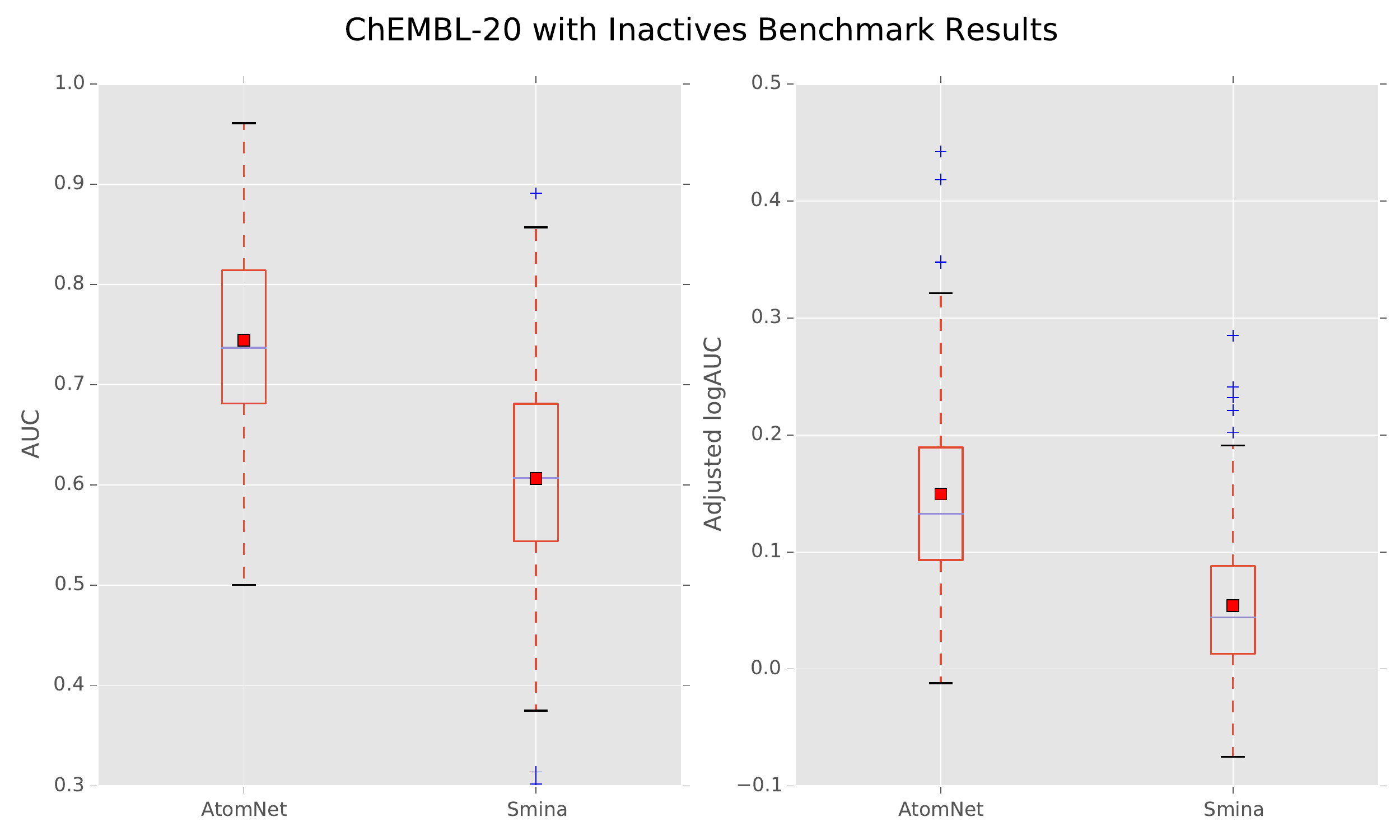}
  \end{center}
  \caption{Distribution of AUC and logAUC values of 149 ChEMBL-20-inactives targets for AtomNet and Smina.}
  \label{fig:chemblInactives}
\end{figure}

\Cref{tab:allRes} and \Cref{fig:chemblPMD,fig:DUDE,fig:chemblInactives} summarize the results across the three different benchmarks.
On each of our four evaluation data sets, AtomNet achieves an order-of-magnitude improvement over Smina at a level of accuracy useful for drug discovery.
\Cref{tab:resCountAUC,tab:resCountLogAUC} summarize the AUC and logAUC results with respect to different performance thresholds.
On the full DUDE set, AtomNet achieves or exceeds 0.9 AUC on 59 targets (or 57.8\%).  
Smina only achieves 0.9 AUC for a single target (\texttt{wee1}), approximately 1\% of the benchmark. 
AtomNet achieves 0.8 or better AUC for 88 targets (86.3\%), while Smina achieves it for 17 targets (16.7\%). 
When we restrict the evaluation to the held-out 30 target subset of DUDE, AtomNet exceeds an AUC of 0.9 and 0.8 for 14 targets (46.7\%) and 22 targets (73.3\%), respectively. 
Smina achieves the same accuracy for 1 target (3.3\%) and 5 targets (16.7\%), respectively. 
AtomNet achieves mean and median AUC of 0.855 and 0.875 on the held-out set compared to 0.7 and 0.694 achieved by Smina, reducing available mean error by 51.6\%. 
As expected, the performance of AtomNet drops slightly for its held-out examples, whereas Smina's performance does not. 

On the ChEMBL-20-PMD dataset, AtomNet achieves an AUC of 0.9 or better for 10 held-out targets (20\% of the set), while Smina achieves it on zero targets. 
When we reduce the standard of accuracy to an AUC of 0.8 or better, AtomNet succeeds on 25 targets (50\%), while Smina only succeeds on 1 target (2\%). 

The third benchmark, which uses inactives instead of property-matched decoys, seems to be more challenging than the other two.
AtomNet predicts with an AUC at or better than 0.9 for 10 targets (6.7\%), while Smina succeeds at zero. 
For meeting or exceeding 0.8 AUC, AtomNet succeeds for 45 targets (30.2\%) and Smina succeeds for 4 (2.7\%).
Although both Atomnet and Smina perform worse than on the previous benchmarks, AtomNet still significantly outperforms Smina with respect to overall and early enrichment performances.
Because this benchmark uses inactives it includes challenging classification cases of structurally similar molecules with different labels~\cite{Hu2012}. 
These cases are excluded from benchmarks using PMD because decoys must be structurally dissimilar in order to presume they can be labelled as inactive.

\begin{figure}[htpb]
  \begin{center}
    \includegraphics[width=\textwidth]{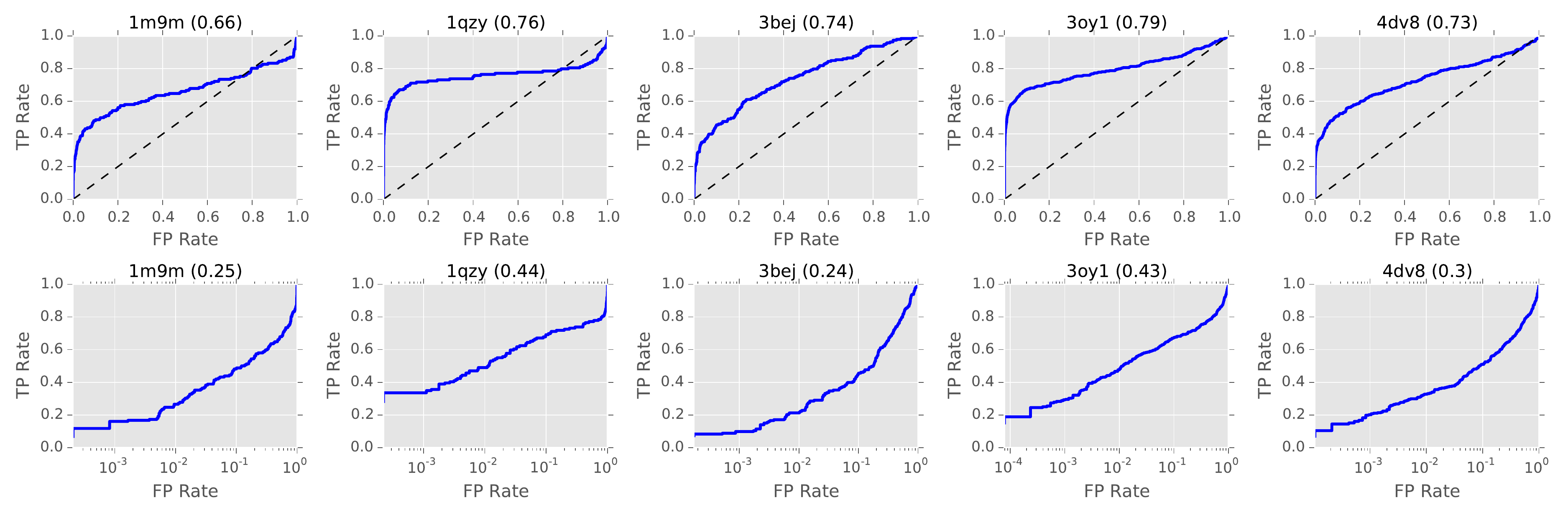}
  \end{center}
  \caption{An illustration of the differences between the AUC and logAUC measurements with respect to the early enrichment.}
  \label{fig:earlyEnrichment}
\end{figure}

Additionally, AtomNet shows good early enrichment performance as indicated by the highly positive logAUC values.
AtomNet outperforms Smina with respect to its early enrichment, achieving a mean logAUC of 0.321 compared to 0.153 of Smina on the DUDE-30 benchmark. 
Visualizing the ROC curves illustrate the difference between the AUC and logAUC measurements with respect to the early enrichment. 
For example, in \Cref{fig:earlyEnrichment} we see that the AUC value for target \textit{1m9m} is 0.66 which may imply mediocre performance. 
However, the early enrichment indicated by the logAUC for that target is 0.25 which suggest that many actives are concentrated at the very top of the rank-ordered results. 
Similarly, target \textit{1qzy} has AUC value of 0.76 but log-scale plot suggest that 35\% of its actives are concentrated at the very top of the rank-order list with logAUC of 0.44.

\begin{table}[!h]
  \centering
  \begin{tabular}{llllll}
    & & \multicolumn{2}{c}{AUC} &  \multicolumn{2}{c}{Adjusted logAUC} \\
    & & Mean & Median & Mean & Median \\ 
    \hline
    \hline
    \multirow{2}{*}{ChEMBL-20 PMD} & AtomNet & 0.781 & 0.792 & 0.317 & 0.328 \\
                                   & Smina & 0.552 & 0.544 & 0.04 & 0.021 \\
    \hline
    \multirow{2}{*}{DUDE-30} & AtomNet &  0.855 & 0.875 & 0.321 & 0.355 \\ 
                          & Smina & 0.7 & 0.694 & 0.153 & 0.139 \\ 
    \hline
    \multirow{2}{*}{DUDE-102} & AtomNet &  0.895 & 0.915 & 0.385 & 0.38 \\  
                          & Smina & 0.696 & 0.707 & 0.138 & 0.132 \\ 
    \hline
    \multirow{2}{*}{ChEMBL-20 inactives} & AtomNet & 0.745 & 0.737 & 0.145 & 0.133 \\ 
                                         & Smina & 0.607 & 0.607 & 0.054 & 0.044 \\ 
    \hline
  \end{tabular}
  \caption{Comparisons of AtomNet and Smina on the DUDE, ChEMBL-20-PMD, and ChEMBL-20-inactives benchmarks. DUDE-30 refers to the held-out set of 30 targets whereas DUDE-102 refers to the full dataset.}
  \label{tab:allRes}
\end{table}

\begin{table}[!h]
  \centering
  \begin{tabular}{lllllll}
    \multicolumn{2}{c}{AUC} & $>0.5$ & $>0.6$ & $>0.7$ & $>0.8$ & $>0.9$ \\ 
    \hline
    \hline
    \multirow{3}{*}{ChEMBL-20 PMD} & AtomNet & 49 & 44 & 36 & 24 & 10  \\ 
                                   & Smina & 38 & 10 & 4 & 1 & 0 \\  
    \hline
    \multirow{3}{*}{DUDE-30} & AtomNet & 30 & 29 & 27 & 22 & 14 \\ 
                             & Smina & 29 & 25 & 14 & 5 & 1 \\ 
    \hline
    \multirow{3}{*}{DUDE-102} & AtomNet & 102  & 101 & 99 & 88 & 59 \\ 
                              & Smina & 96 & 84 & 53 & 17 & 1 \\ 
    \hline
    \multirow{3}{*}{ChEMBL-20 inactives} & AtomNet & 149 & 136 & 105 & 45 & 10 \\ 
                                         & Smina & 129 & 81 & 31 & 4 & 0 \\ 
    \hline
  \end{tabular}
  \caption{The number of targets on which AtomNet and Smina exceed given AUC thresholds.  For example, on the CHEMBL-20 PMD set, AtomNet achieves an AUC of 0.8 or better for 24 targets (out of 50 possible targets).  ChEMBL-20 PMD contains 50 targets, DUDE-30 contains 30 targets, DUDE-102 contains 102 targets, and ChEMBL-20 inactives contains 149 targets.}
  \label{tab:resCountAUC}
\end{table}

\begin{table}[!h]
  \centering
  \begin{tabular}{lllllll}
    \multicolumn{2}{c}{Adjusted-LogAUC} & $>0.0$ & $>0.1$ & $>0.2$ & $>0.3$ & $>0.4$ \\ 
    \hline
    \hline
    \multirow{2}{*}{ChEMBL-20 PMD} & AtomNet & 49 & 44 & 36 & 27 & 20 \\
                                   & Smina & 35 & 8 & 2 & 1 & 0 \\
    \hline
    \multirow{2}{*}{DUDE-30} & AtomNet & 30 & 27 & 22 & 17 & 10 \\ 
                             & Smina & 29 & 19 & 8 & 2 & 1 \\ 
    \hline
    \multirow{2}{*}{DUDE-102} & AtomNet & 102  & 99 & 88 & 69 & 43 \\ 
                              & Smina & 94 & 65 & 28 & 5 & 1 \\ 
    \hline
    \multirow{2}{*}{ChEMBL-20 inactives} & AtomNet & 147 & 107 & 36 & 10 & 2 \\ 
                                         & Smina & 123 & 35 & 5 & 0 & 0 \\ 
    \hline
  \end{tabular}
  \caption{The number of targets on which AtomNet and Smina exceed given adjusted-logAUC thresholds.  For example, on the CHEMBL-20 PMD set, AtomNet achieves an adjusted-logAUC of 0.3 or better for 27 targets (out of 50 possible targets).  ChEMBL-20 PMD contains 50 targets, DUDE-30 contains 30 targets, DUDE-102 contains 102 targets, and ChEMBL-20 inactives contains 149 targets.}
  \label{tab:resCountLogAUC}
\end{table}


\section{Discussion}
\begin{figure}[htpb]
  \begin{center}
    \includegraphics[width=0.49\textwidth]{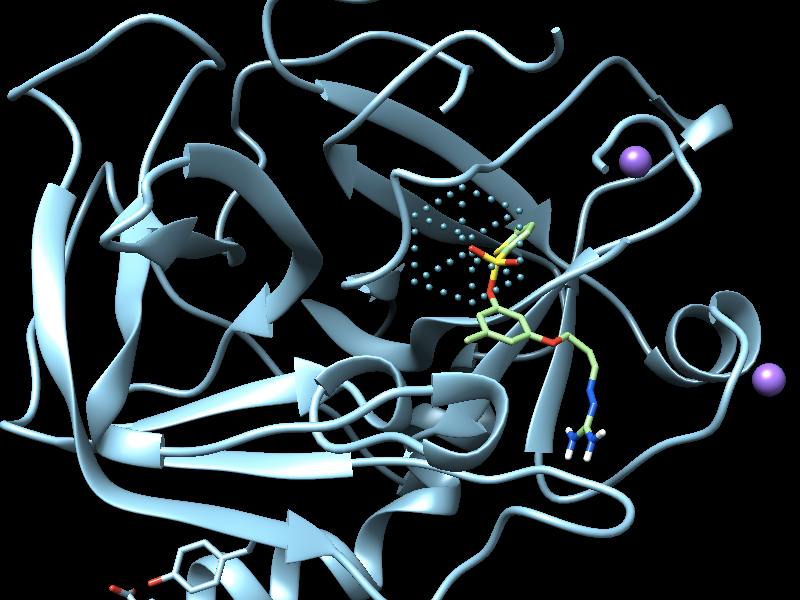}
    \includegraphics[width=0.49\textwidth]{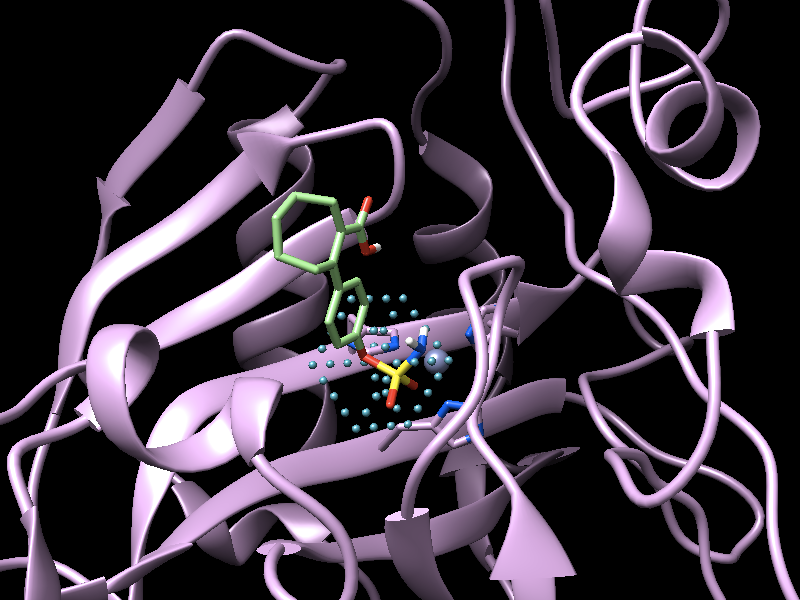}
  \end{center}
  \caption{Sulfonyl/sulfonamide detection with autonomously trained convolutional filters.}
  \label{fig:Filters}
\end{figure}

\paragraph{Filter visualization}
Convolutional layers consist of multiple different filters that learn to identify specific locally-related features by repeatedly applying these filters across the receptive field. 
When dealing with images, one can visualize these filters to verify that the model is capable of learning relevant features.
For example, Krizhevsky~{\em et~al.}~\cite{Krizhevsky2012} demonstrated that filters in the first convolutional layer of their model could detect lines, edges, and color gradients. 
In our case, however, we can not easily visualize the filters because: (i) the filters are 3-dimensional, and (ii) the input channels are discrete. 
For example, two close RGB values will result with two similar colors but carbon is not closer to nitrogen than to oxygen. That is, similar values do not imply similar functionalities.
To overcome these limitations we take an indirect approach. 
Instead of directly visualizing filters in order to understand their specialization, we apply filters to input data and examine the location where they maximally fire.
Using this technique we were able to map filters to chemical functions.
For example, \Cref{fig:Filters} illustrate the 3D locations at which a particular filter from our first convolutional layer fires.
Visual inspection of the locations at which that filter is active reveals that this filter specializes as a sulfonyl/sulfonamide detector.
This demonstrates the ability of the model to learn complex chemical features from simpler ones.
In this case, the filter has inferred a meaningful spatial arrangement of input atom types without any chemical prior knowledge.

\paragraph{Comparison to other structure-based methods}
The aim of this paper is to present a novel application of deep convolutional neural networks to bioactivity predictions, rather than reporting head-to-head comparisons to other structure-based methods.
In order to put results in context, we used the popular program Smina as a baseline point of reference.
Smina has practical advantages: it is fast, free, and under active development, so we find it suitable for analyzing large benchmarks in a timely and cost-efficient manner.
Nevertheless, using published work, we can provide broader context by comparing AtomNet to other commercial docking algorithms reported in the literature.
The most common benchmark is DUDE~\cite{Mysinger2012} which, like Smina, is publicly available and widely used. 
We therefore present the following comparisons to previously described results:
\begin{compactitem}
\item Gabel~{\em et~al.}~\cite{Gabel2014} evaluated Surflex-Dock~\cite{Spitzer2012} on a representative set of 10 targets from the DUDE. 
The median AUC of Surflex-Dock was 0.76 compared to 0.93 achieved by AtomNet.
\item Coleman~{\em et~al.}~\cite{Coleman2014} evaluated DOCK3.7~\cite{Coleman2013} in a fully automated manner over the whole DUDE benchmark. 
They achieved mean AUC of 0.696 and logAUC of 0.174 compared to our AUC of 0.895 and logAUC of 0.385.
\item Allen~{\em et~al.} reported mean AUC of 0.72 on 5 DUDE targets using Dock6.7~\cite{Allen2015} compared to AUC of 0.852 by AtomNet.
\end{compactitem}

\section{Conclusion}
We presented AtomNet, the first structure-based deep convolutional neural network, designed to predict the bioactivity of small molecules for drug discovery applications.  
The locally-constrained deep convolutional architecture allows the system to model the complex, non-linear phenomenon of molecular binding by hierarchically composing proximate basic chemical features into more intricate ones. 
By incorporating structural target information AtomNet can predict new active molecules even for targets with no previously known modulators.  
AtomNet shows outstanding results on a widely used structure-based benchmark achieving an AUC greater than 0.9 on 57.8\% of the targets in the DUDE benchmark, far surpassing previous docking methods.

\bibliographystyle{ieeetr}
\bibliography{AtomNet.bib}

\end{document}